# Lexical Simplification using multi level and modular approach

**Nikita Katyal**
nikita18katyal@gmail.com

**Pawan Kumar Rajpoot**
pawan.rajpoot2411@gmail.com

## Abstract

Text Simplification is an ongoing problem in Natural Language Processing, solution to which has varied implications. In conjunction with the TSAR-2022 Workshop @EMNLP2022 Lexical Simplification is the process of reducing the lexical complexity of a text by replacing difficult words with easier to read (or understand) expressions while preserving the original information and meaning. This paper explains the work done by our team "teamPN" for English sub task. We created a modular pipeline which combines modern day transformers based models with traditional NLP methods like paraphrasing and verb sense disambiguation. We created a multi level and modular pipeline where the target text is treated according to its semantics(Part of Speech Tag). Pipeline is multi level as we utilize multiple source models to find potential candidates for replacement, It is modular as we can switch the source models and their weight-age in the final re-ranking.

## 1 Introduction

As per TSAR-2022 workshop shared task the problem definition is: "Given a sentence containing a complex word, systems should return an ordered list of simpler valid substitutes for the complex word in its original context." Some examples are shown in figure 1. The English data-set consists of 373 sentences, with 1 complex word per sentence. No training data was provided and the teams were free to create supervised or unsupervised model. We found that majority of the complex words were ambiguous verbs. Hence semantic interpretation and separate treatment of verbs is essential to solve complexity of verbal behaviors. Verbs are generally more ambiguous in the senses which they are used when compared to other Part of Speech tags. We based our whole idea on this assumption and we treated verbs with an additional module apart from usual ones. Other than this special module we had 3 standard ones. One where we used Para Phrase Data Base for directly finding synonyms for complex words. Second we used new age Transformer based models(Language Modeling) where we masked the complex word and predicted its alternatives. Third we used Knowledge Graph based fine grained Entity recognition and hence used predicted entities synonym nodes in KG as potential candidates. Finally we aggregated and re-ranked all candidates again using Transformers based model.

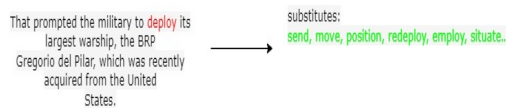

Figure 1: TSAR task sample.

## 2 Approach

We parse the sentence using spacy and run different sets of modules for verb, noun and adjectives respectively. Our modules are explained in in detail as follows. See Algorithm 11 for pseudo code of the pipeline.

### 2.1 Potential Candidate Collection

#### 2.1.1 Verb Sense Disambiguation

Verbnet is a lexicon which is an extension to Levin's original Verb classifications(Levin (1993)) in 1993. Semantically similar verbs are placed in same class. We use Verbnet 3.1 (Schuler (2005)) to ground the verb and get possible classes. For class prediction we do not rely on traditional VSD work Abend et al. (2008) , Dligach and Palmer (2008) and Kawahara and Palmer (2014) as the data which is used in model training is WSJ data(Loper et al. (2007)) which is heavily biased. For instance the verb "rise" has 6 possible classes in verbnet, but in WSJ data 93 percent of the examples have "rise" related to "calibration", as in "Stocks rise from 10 to 12". Instead we rely on modern day Transformer based Language Models. We first mask the target word and use FitBERT(Havens and Stal (2019)) to rank the top possible words among all possible classes member verbs. We choose the verbnet class with maximum representations in top k predicted words. Once the class is fixed we return the class members as potential candidates.

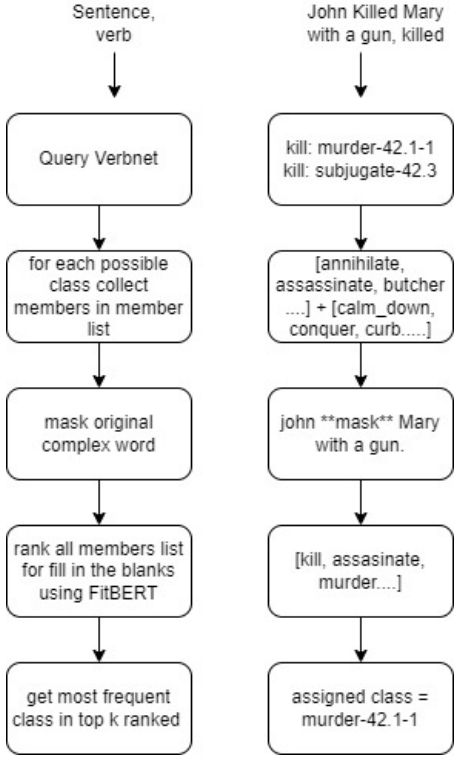

Figure 2: Verb Sense Disambiguation Module.

### 2.1.2 Paraphrase Data Base

For Nouns and adjectives we directly query PPDB(Ganitkevitch et al. (2013)) and return the retrieved result list as potential candidates. We use lexical version and small size dictionary as it contains the high quality paraphrases.

### 2.1.3 Distil BERT

This module is common for all 3 POS types viz verbs, nouns and adjectives. We mask the complex word in the context and then use distil BERT(Sanh et al. (2019)) model to predict the words(fill-mask pipeline) then return the result list as the potential candidates. Due to computational resource restrictions we were not able to use high end Transformer models.

### 2.1.4 Knowledge Graph

We use Multi Modal Knowledge Graph VisualSem (Alberts et al. (2020)) to do text entity extraction from the sentence for target complex word. Then we use the "synonym" relation to get the entity nodes which are then collected in a list and returned as potential candidates. For entity extraction, CLIP textual embedding(Radford et al. (2021)) were used as defined in original Visualsem paper.

## 2.2 Aggregation and re-ranking

See Table 1 for usage of modules as per POS tags. Once all potential candidate list is created first we combine all together, then we adjust all the inflections similar to that of original complex word. Then we again use FitBERT(Havens and Stal (2019)) to rank the combined candidates. For the submissions we used 5 top words.

---
**Algorithm 1** teamPN: Text Simplification
---
**Require:** m1 = vsdModule
**Require:** m2 = PPDBModule
**Require:** m3 = distilBertModule
**Require:** m4 = kgModule
  **for** each sentence and complexWord **do**
    pos = getPos(complexWord)
    **if** pos == verb **then**
      candidates = m1 + m2 + m3 + m4
    **end if**
    **if** pos == noun **then**
      candidates = m2 + m3 + m4
    **end if**
    **if** pos == adj **then**
      candidates = m2 + m3
    **end if**
    candidates = fixInflection(candidates)
    rankCandidates = rerankUsingFitBERT
  **end for**
---

| POS/Module | VSD | PPDB | distil BERT | KG |
|---|---|---|---|---|
| VERB | Y | Y | Y | N |
| NOUN | N | Y | Y | Y |
| ADJ | N | Y | Y | N |

Table 1: Use of Candidate collection modules as per part of Speech of complex word.

## 3 Results

As per TSAR definition (Štajner et al. (2022)) The evaluation metrics to be applied in the TSAR-2022

Shared Task are the following:
MAP@K (Mean Average Precision @ K): K=1,3,5,10. The MAP@K metric is used to check whether the predicted word can be matched (relevant) or not matched (irrelevant) against the set of the gold-standard annotations for evaluation. MAP@K for Lexical Simplification evaluates the following aspects: 1) are the predicted substitutes relevant?, and 2) are the predicted substitutes at the top positions?

Potential@K: K=1,3,5,10. The percentage of instances for which at least one of the substitutions predicted is present in the set of gold annotations.

Accuracy@K@top1: K=1,2,3. The ratio of instances where at least one of the K top predicted candidates matches the most frequently suggested synonym/s in the gold list of annotated candidates.

We stand 12th, on the official results Saggion et al. (2022) link of TSAR-2022 shared task. We outperform one of the baseline models TUNER (Štajner et al. (2022)). See Table 2 for our scores.

| Metric | Score |
|---|---|
| ACC@1 | 0.4664 |
| ACC@1@Top1 | 0.1823 |
| ACC@2@Top1 | 0.3056 |
| ACC@3@Top1 | 0.3378 |
| MAP@3 | 0.2743 |
| MAP@5 | 0.195 |
| MAP@10 | 0.0975 |
| Potential@3 | 0.6729 |
| Potential@5 | 0.7506 |
| Potential@10 | 0.7506 |

Table 2: Our scores for TSAR 2022 shared tasks

## 4 Conclusion and Future Work

We presented a novel approach where we combine power of new age transformer models with traditional NLP work. Our work was restricted by computing resources. We would further like to improve on our modules using more modules built out from complex transformers. Also apart from PPDB we did not work with any other synonym dictionaries, adding more open source dictionary modules will bring on more variety. All of our code and documentation is available at our git link

## Acknowledgements

We would like to acknowledge TSAR organizing committee and EMNLP 2022 for their support and also organizing the workshop event.